\documentclass[letterpaper]{article} 
\usepackage[preprint,nonatbib]{neurips_2019}  
\usepackage{url}  
\usepackage{tikz}
\usetikzlibrary{arrows}
\usepackage{xspace}
\usepackage{latexsym}
\usepackage{amsfonts}
\usepackage{amsmath}
\usepackage{amssymb}
\usepackage{color}
\usepackage{colortbl}
\usepackage{epsfig}
\usepackage{graphicx}
\usepackage{pbox}
\usepackage{paralist}
\usepackage{enumerate}
\usepackage[color,matrix,arrow,all]{xy}
\usepackage{comment}
\usepackage{booktabs}
\usepackage{balance}
\usepackage{stmaryrd}
\usepackage{pifont}
\usepackage{hhline}
\usepackage{listings}
\usepackage{array}
\usepackage[flushleft]{threeparttable}
\usepackage{subcaption}
\usetikzlibrary{arrows}
\usepackage{enumitem}

\DeclareMathAlphabet{\pazocal}{OMS}{zplm}{m}{n}

\usepackage{array}
\newcolumntype{M}{>{\begin{varwidth}{2cm}}l<{\end{varwidth}}}
\newcolumntype{L}[1]{>{\raggedright\let\newline\\\arraybackslash\hspace{0pt}}m{#1}}
\newcolumntype{C}[1]{>{\centering\let\newline\\\arraybackslash\hspace{0pt}}m{#1}}
\newcolumntype{R}[1]{>{\raggedleft\let\newline\\\arraybackslash\hspace{0pt}}m{#1}}

\newcommand{\ie}{{\em i.e.,}\xspace}
\newcommand{\eg}{{\em e.g.,}\xspace}

 \usepackage[framemethod=TikZ]{mdframed}
\usetikzlibrary{shadows}
\mdfdefinestyle{myframe}{%
    linecolor=black,
    outerlinewidth=0.6pt,
    roundcorner=10pt,
    innertopmargin=10pt,
    innerbottommargin=10pt,
    innerrightmargin=10pt,
    innerleftmargin=10pt,
    skipabove=0.6\baselineskip,
    }

\title{Benchmarking Regression Methods: A comparison with CGAN}

\author{
	Karan Aggarwal\\
	University of Minnesota\\
	\texttt{aggar081@umn.edu} \\
	\And
	Matthieu Kirchmeyer\\
	Criteo AI Lab \\
	\texttt{m.kirchmeyer@criteo.com} \\
	\AND
	Pranjul Yadav \\
	Criteo AI Lab  \\
	\texttt{p.yadav@criteo.com} \\
	\And
	S. Sathiya Keerthi\\
	Criteo AI Lab  \\
	\texttt{s.selvaraj@criteo.com} \\
	\And
	Patrick Gallinari\\
	Criteo AI Lab  \\
	\texttt{p.gallinari@criteo.com} \\
}
\begin{document}
	
	\maketitle
	
	\begin{abstract}
		In recent years, impressive progress has been made in the design of implicit probabilistic models via Generative Adversarial Networks (GAN) and its extension, the Conditional GAN (CGAN). Excellent solutions have been demonstrated mostly in image processing applications which involve large, continuous output spaces. There is almost no application of these powerful tools to problems having small dimensional output spaces. Regression problems involving the inductive learning of a map, $y=f(x,z)$, $z$ denoting noise, $f:\mathbb{R}^n\times \mathbb{R}^k \rightarrow \mathbb{R}^m$, with $m$ small (e.g., $m=1$ or just a few) is one good case in point. The standard approach to solve regression problems is to probabilistically model the output $y$ as the sum of a mean function $m(x)$ and a noise term $z$; it is also usual to take the noise to be a Gaussian. These are done for convenience sake so that the likelihood of observed data is expressible in closed form. In the real world, on the other hand, stochasticity of the output is usually caused by missing or noisy input variables. Such a real world situation is best represented using an implicit model in which an extra noise vector, $z$ is included with $x$ as input. CGAN is naturally suited to design such implicit models. This paper makes the first step in this direction and compares the existing regression methods with CGAN. 

We notice however, that the existing methods like mixture density networks (MDN) and XGBoost do quite well compared to CGAN in terms of likelihood and mean absolute error, respectively. Both these methods are comparatively easier to train than CGANs. CGANs need more innovation to have a comparable modeling and ease-of-training with respect to the existing regression solvers. In summary, for modeling uncertainty MDNs are better while XGBoost is better for the cases where accurate prediction is more important. 
	\end{abstract}
	
	\section{Introduction}

\def\to{\rightarrow}
\def\eps{z}
\def\N{{\cal{N}}}

Regression is an important problem in statistics and machine learning~\cite{bishop2006pattern}. In regression, the true output ($y\in \mathbb{R}^m$) is a continuous and stochastic function of the input ($x\in \mathbb{R}^n$): 
\begin{equation}
y = f(x,\eps) \;\; \mbox{where} \;\; \eps\in \mathbb{R}^k \;\; \mbox{is the noise vector.} \label{eq1}
\end{equation}
Regression methods attempt to model $f$ by induction using a training data of many $(x,y)$ pairs collected from the real world. Most of these methods~\cite{bishop2006pattern} model the noise in an additive manner, \ie
\begin{equation}
y = \hat{f}(x) + \eps \label{eq2}
\end{equation}
where $\eps$ is an additive unimodal noise with parameters; for example, $\eps \sim \N(0,\Sigma)$, the zero-mean Gaussian with covariance matrix $\Sigma$. This is usually done for convenience, to write a closed form expression for the conditional density, $p(y|x)$. Regression using Gaussian processes (GPs)~\cite{williams1996gaussian} is a great example of a commonly used regression method that uses this additive noise modeling. 

There are also some extended types of additive noise modeling: heteroscedastic regression, in which the noise $\eps$ in (\ref{eq2}) is chosen as a function of $x$; regression with multi-modal posteriors, where $p(y|x)$ is a multi-modal distribution. Typically, a separate set of methods is developed for each such type of modeling, for example, the extended heteroscedastic GP method~\cite{le2005heteroscedastic}.

All the above mentioned models are special cases of (\ref{eq1}). Also, in real world systems, how noise enters the true $y$-generation process is generally unknown, and so, ideally it is best to leave it to a non-parametric method to form $f$. For instance, in many situations, stochasticity of $y$ arises because several input variables are unknown and affect $y$ jointly in a non-linear fashion together with the known variables; in general, the dimensionality of the set of unknown variables is also not known. 

\begin{figure}[t]
\centering
    \begin{subfigure}[b]{0.6\textwidth}
           \centering
           \includegraphics[width=\textwidth]{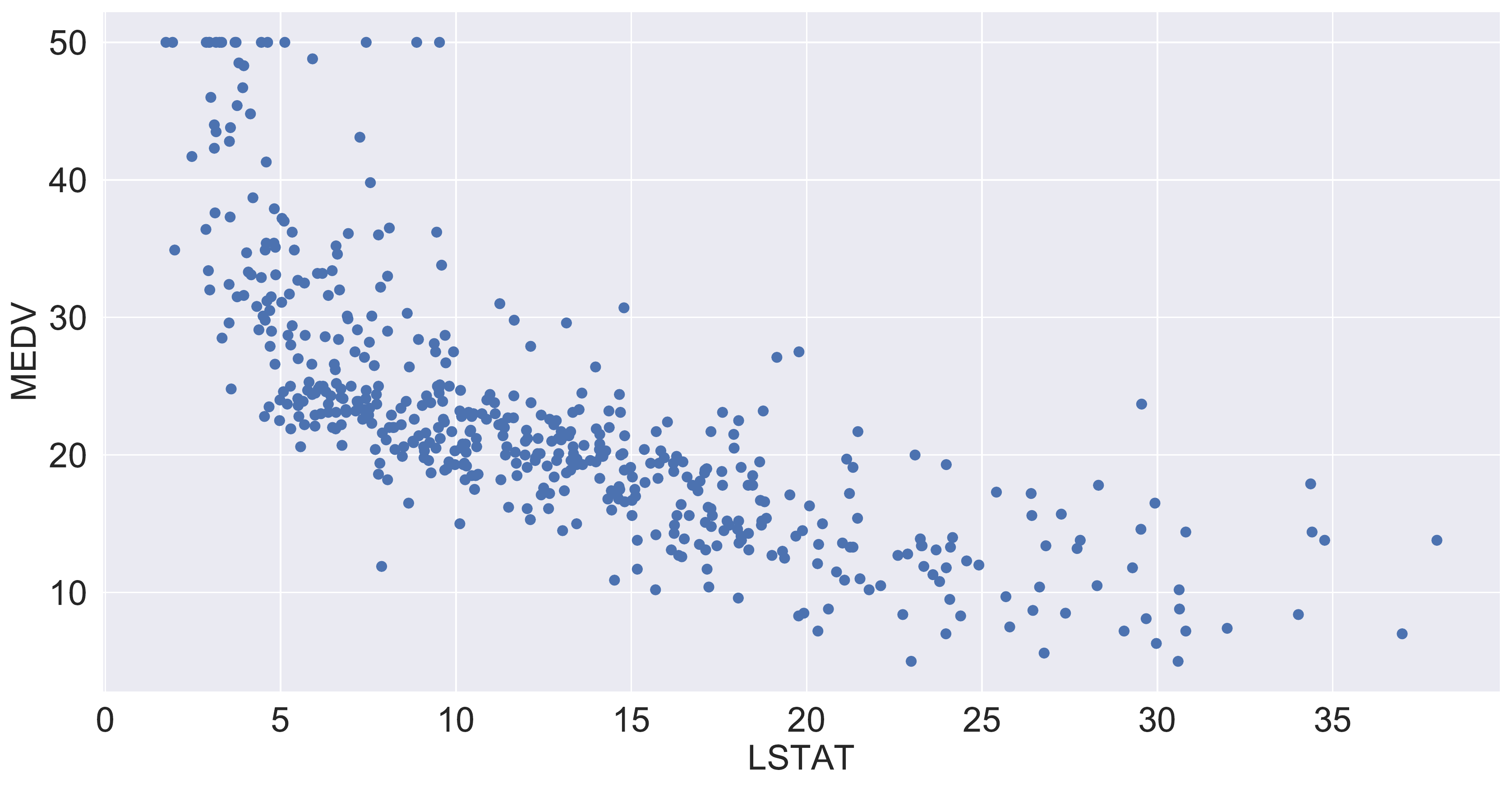}
            \caption{}
            \label{fig:a}
    \end{subfigure}
    \begin{subfigure}[b]{0.32\textwidth}
           \centering
           \includegraphics[width=\textwidth]{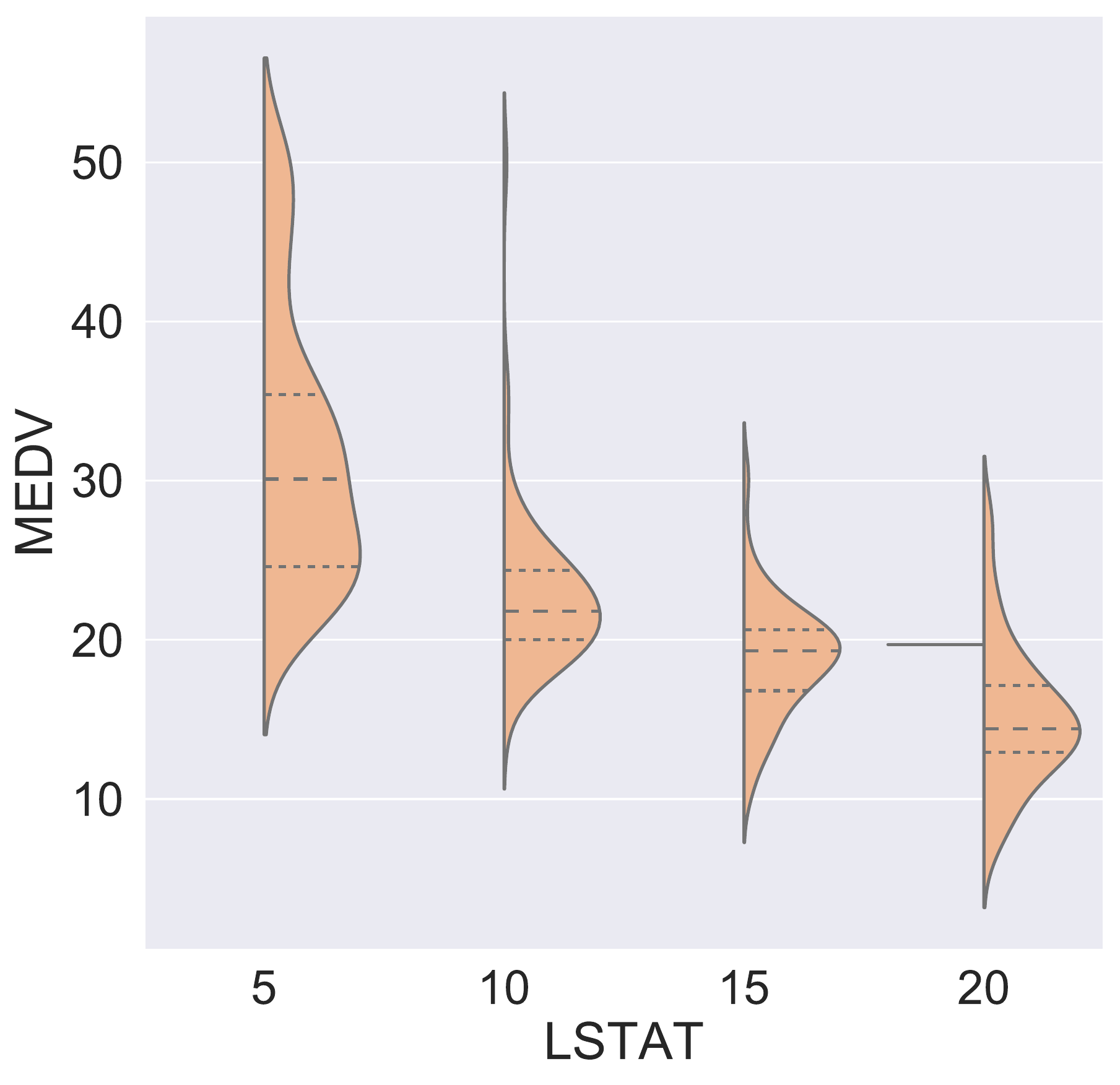}
            \caption{}
            \label{fig:a}
    \end{subfigure}
\caption{(a) Boston housing dataset with one input variable, $x=LSTAT$ and output variable $y=MEDV$. (b) The smoothened histogram of $MEDV$ values around $LSTAT = 5, 10, 15, 20$.
}
\label{fig:boston}
\end{figure}

To illustrate, consider the {\tt Boston Housing} dataset\footnote{\url{https://www.cs.toronto.edu/~delve/data/boston/bostonDetail.html}} with thirteen input variables and one target variable, $y=MEDV$. Suppose only one input variable, $LSTAT$ is available, and we want to model $p(MDEV|LSTAT)$. Figure~\ref{fig:boston} shows the data graphically; it also shows the (smoothened) histogram of $MDEV$ values around $LSTAT = 5, 10, 15, 20$. Clearly, the target noise patterns are quite different at different $LSTAT$ values. This is due to the different effects of the missing input variables at different LSTAT values. Clearly it is better to model such a function via (\ref{eq1}) where $z$ brings the effect of the missing variables. It is reasonable to take $z$ to be a known distribution, e.g., standard multivariate Gaussian, and leave all required transformations to be directly modeled by a non-parametric model $f$.


The main difficulty with (\ref{eq1}) is that (\ref{eq1}) is an implicit probabilistic model and hence, forming and working with the density function, $p(y|x)$ is hard. In recent years, impressive progress has been made in Conditional Generative Adversarial Networks (CGANs)~\cite{mirza2014conditional} precisely to handle this difficulty. CGAN uses an auxiliary non-parametric function (e.g., DNN) to model the loss function (e.g., Jenson-Shannon or KL divergence) between the implicit probabilistic output model in (\ref{eq1}) and the true $p(y|x)$ represented by the training samples.

However, CGAN has been mainly applied to domains such as images~\cite{antipov2017face} where $m$, the dimension of $y$ is large (several thousands) and the noise dimension, $k$ is much smaller (ten to twenty). Little effort\footnote{Chapfuwa et al.~\cite{chapfuwa2018adversarial} is a special use of CGAN with survival analysis for time-to-event modeling.} has been made to explore the use of CGAN for solving regression problems where $k \geq m$. This paper aims to fill this gap. We implement a basic CGAN set up for regression (section~\ref{sec:setup}) without any extensive tuning of hyperparameters\footnote{Our CGAN experiments can be reproduced using our code from: \url{https://github.com/kaggarwal/ganRegression}.}. We take state-of-the-art regression methods (section~\ref{sec:setup}) such as Gaussian processes (GPs), Deep neural nets (DNNs) and XGboost as the baseline methods to compare and evaluate CGAN. Using synthetic datasets we show (sections \ref{sec:linear}-\ref{sec:real-world}) that CGAN possesses a natural and better ability to model complex noise forms. We also conduct experiments on seven real world datasets to demonstrate that even our straight-forward adaptation of CGAN is competitive with state-of-the art methods for regression. We point to several advantages and improvements of CGAN for regression, in section~\ref{sec:conc}. Therefore, further research on the application of CGANs for regression, focused on improving its performance, efficiency and robustness, is very worthwhile.


	\section{Related Work}
	
Building models that generate distributions similar to underlying data has been a defining topic of machine learning research. Most generative models assume a specific form (such as a nonlinear function for the mean with Gaussian additive noise) for the  probability density function and maximize the likelihood of data. The nonlinear function can be formed using kernel methods, deep nets, boosted trees~\cite{xgboost}, etc. One can also apply a Bayesian approach over such a design to get improved likelihood properties - a good example is Gaussian processes (GPs)~\cite{williams1996gaussian}. For more complicated noise forms such as hetero-scedastic noise, specialized models have to be built~\cite{le2005heteroscedastic}.

Generative adversarial network (GAN)~\cite{goodfellow2014generative} and its conditional variant, CGAN~\cite{mirza2014conditional} have demonstrated a great ability to generate realistic samples of an underlying distribution using implicit probability modeling without any specific assumption about the nature of the probability density function. However, most of this research has focused on the image domain that is characterized by a large structured output space. Little attention has been paid to apply GANs to other classes of problems such as regression that typically have a much smaller output space. 

In this study, we generate a continuous output distribution implicitly using $y=f(x,z)$ where $z$ has a specified distribution and $f$ is modeled using deep nets; we use the CGAN~\cite{mirza2014conditional} approach to design this model. Chapfuwa et al.~\cite{chapfuwa2018adversarial} specifically use CGANs with survival analysis for time-to-event modeling.

	\section{Experimental Setup and Implementation}
	\label{sec:setup}


We use the standard CGAN~\cite{mirza2014conditional} formulation in which the discriminator network corresponds to minimizing the Jenson-Shannon divergence between the true and modeled $p(y|x)$ distributions. Alternatively, one could use Wasserstein GAN~\cite{arjovsky2017wasserstein}, or f-GAN ideas for minimizing KL divergence~\cite{nowozin2016f}. Note that each of these models optimizes a different loss function, which could impact the performance. 


We report two metrics with each experiment: Negative Log Predictive Density (NLPD) and Mean Absolute Error (MAE). NLPD is taken as the main metric of interest since, for many real world regression problems, studying the goodness of the estimate of uncertainty is of great interest. 

\paragraph{NLPD computed using Parzen windows}
Since CGAN models $p(y|x)$ implicitly, it can only generate samples of $y$ for each given $x$. In the GAN literature, Parzen windows has been used~\cite{bengio2013better,goodfellow2014generative} to approximate $p(y|x)$ using such samples and then use it for evaluating NLPD; we do the same in this paper. We generate 100 samples of $y$ for each $x$ to build the Parzen windows distribution. Note that probability density can take unlimited positive values, and hence, NLPD can be negative.



\paragraph{MAE}
Given a test dataset, $\{(x_i,y_i)\}_{i=1}^N$, MAE for a regressor $f$ is defined as $\dfrac{1}{N} \sum_{i=1}^N |y_i - \widehat{y}_i|$ where $\widehat{y}_i$ is a single central value returned by the regressor at $x=x_i$. For regressors that generate a distribution $p(y|x_i)$, we define $\widehat{y}_i$ to be the median of $p(y|x_i)$; note that median is the central estimate that minimizes mean absolute error. For GP, $p(y|x_i)$ is a Gaussian, and so $\widehat{y}_i$ is taken to be the mean of $p(y|x_i)$ as returned by the model. For CGAN, we generate a sample of 100 $y$ values for the given $x_i$ and take the median of the sample points to be $\widehat{y}_i$. 

\paragraph{Confidence and uncertainty when reporting metrics}
Given the underlying uncertainty in the evaluation of the CGAN method (dependent on the sample drawn from $p(y|x)$), we report the mean over $10$ evaluation runs. We found the standard deviation of the metrics over these different samples to be small relative to the mean values ($< 10^{-3}$).  In later sections, when we report metrics, they are mean values from 10 runs; also, when we mention a value in \textbf{bold}, it will mean that the method whose metric is reported in bold is statistically significantly better than the other model on that metric.



\paragraph{Datasets} We use two types of datasets: synthetic and real world.
We generate four datasets with increasing complexity in noise and the nature of the $f$ function in (\ref{eq1}). 
Description of these datasets and the results on them are given in sections~\ref{sec:linear}-~\ref{sec:multi-response}. We used seven real world datasets (see section~\ref{sec:real-world} for results) taken from \url{http://www.dcc.fc.up.pt/~ltorgo/Regression/DataSets.html}. 

\paragraph{Preprocessing} A different preprocessing strategy was used on synthetic data and real world data. On real world experiments, input features were scaled to have zero mean and unit variance. 
On the synthetic datasets, we do not scale features as they already have a reasonably normalized range of inputs and outputs.

\paragraph{Baseline Models} Most practitioners employ GPs, DNNs and Boosted trees (XGboost) as the state-of-the-art regression solvers.
For GP, we use the GPy package~\cite{gpy2014} with automatic hyperparameter optimization. We tried both, the radial basis function (RBF) kernel and the Rational quadratic (Quad) kernel.
We use \texttt{Keras}~\cite{chollet2015} for DNNs.
For XGboost we use XGboost package~\cite{xgboost}. 
All these models are based on the additive Gaussian noise model. While GPs provide NLPD values directly, we use Gaussian likelihood assumption for calculating the NLPD values for DNN and XGboost.

We also use Mixture Density Networks (MDN)~\cite{bishop1994mixture} as a baseline, since they explicitly model $y$ as a mixture of Gaussians, hence giving them ability to model very complex noise forms. We use similar architecture as DNNs with ReLu activation units.


\paragraph{Dataset statistics} We split our datasets into train, validation, and test sets.
On synthetic datasets, each set is taken to be of size $1000$. On real world datasets we generate the validation set by picking 25\% of the training set. Dataset statistics can be checked at \url{https://www.cs.toronto.edu/~delve/data/datasets.html}.

\paragraph{Hyperparameter tuning} Hyperparameter tuning was done using the validation set. Hyperparameters were tuned to minimize NLPD since that is our main metric of interest. 


We use DNN for CGAN. Following are the details of CGAN architecture used in our experiments. We use one base architecture for all our experiments. The \emph{Generator} uses a six layered network. We separately feed input $x$ and noise $z$ through a three-layered MLP and concatenate the output representations to a three-layered network. Except for the final layer that employs linear activation, we use exponential linear unit (ELU)~\cite{clevert2015fast} as activation function. Addition of a direct connection from $z$ to the final linear unit can help model additive noise but was not necessary. We generally found ELU to work better than activations such as ReLU, leaky-ReLU, or tanh.
The \emph{Discriminator} uses a four-layered network. It feeds inputs $x$ and $y$ through a single non-linear layer, and then passes the concatenated outputs through a three-layered MLP whose final layer uses sigmoid activation.


The above architecture worked well on synthetic datasets. The following choice of number of neurons per layer worked well in general: 40 neurons for complex noise datasets (\texttt{heteroscedastic} and \texttt{multi-modal}) and 15 for simple additive noise datasets (\texttt{linear} and \texttt{sinus}). For real world datasets we increase the number of layers and neurons. For real world datasets, we use a seven-layered network, with one layer of size 100 for $x$ and noise $z$, whose outputs are concatenated and passed on to another set of six-layers of size 50. We use same architecture for DNN baseline but using ReLU activation, since that worked better in our experiments.

We use the Adam optimizer and tune it with learning rates of $\{10^{-2}, 10^{-3}, 10^{-4}\}$. We found the learning rates of $10^{-4}$ for generator and $10^{-3}$ for discriminator works uniformly well. Improvements are possible with a decay of $10^{-3}$ on generator's learning rate. 

The number of epochs was fixed to $2000$ on synthetic experiments and $500$ on real world datasets. The ratio of training steps of the discriminator over that of the generator was set to 1. Loss curves are stable and converge to the cross entropy of a random discriminator prediction. Batch size of 100 was used in our experiments. The dimension of the noise $z$ was fixed to one on all experiments (including synthetic and real-world). We increase this dimension in the studies of Section \ref{sec:multi-response} and \ref{sec:real-world} to analyze the effect of higher dimnensional noise. Experiments were run on a machine with 16GB RAM and 2.7 GHz processor speed. No significant speedup was noticed when using a Tesla M40 24GB GPU card. 

\paragraph{Visualization of experiments}
Visualization plots are used on synthetic datasets in sections \ref{sec:linear}-\ref{sec:multi-response} to compare CGAN against baseline methods. To avoid clutter, among the baseline methods we only use GP (RBF). 
We take key $x$ values and plot the conditional probability density using a Gaussian kernel density estimator. To do this, at each $x_i$ we form a sample of size 200 from $p(y|x_i)$ for each method (CGAN or any others). Due to the use of the kernel density estimator, on some plots we may not see perfect Gaussian distributions for GP. To further investigate distributions being generated (predicted) by these methods, we also plot generated samples from CGAN and GP.


 
 \begin{figure*}[b!]
    \centering
    \begin{subfigure}[b]{0.3\textwidth}
           \centering
           \includegraphics[width=\textwidth]{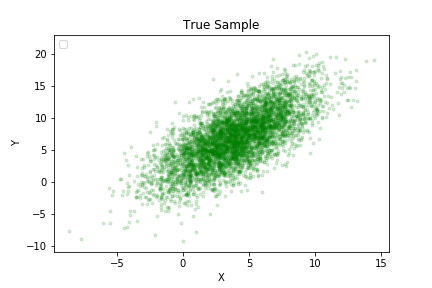}
            \caption{}
            \label{fig:a}
    \end{subfigure}
    \begin{subfigure}[b]{0.3\textwidth}
            \centering
            \includegraphics[width=\textwidth]{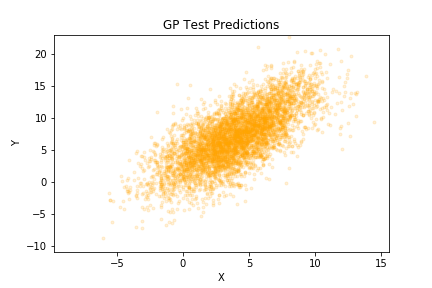}
            \caption{}
            \label{fig:b}
    \end{subfigure}
\begin{subfigure}[b]{0.3\textwidth}
            \centering
            \includegraphics[width=\textwidth]{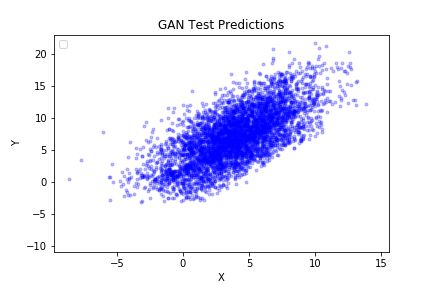}
            \caption{}
            \label{fig:c}
    \end{subfigure}
\newline\vspace{-1em}
    
      \begin{minipage}{\textwidth}
 \begin{minipage}[b]{0.6\textwidth}
    \begin{subfigure}[b]{0.48\textwidth}
            \centering
            \includegraphics[width=\textwidth]{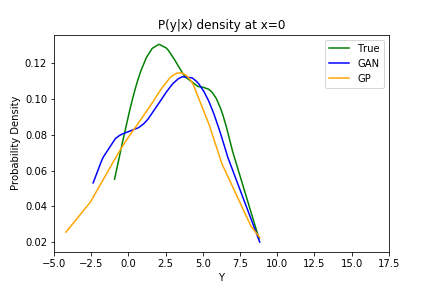}
            \caption{}
            \label{fig:b}
    \end{subfigure}
\begin{subfigure}[b]{0.48\textwidth}
            \centering
            \includegraphics[width=\textwidth]{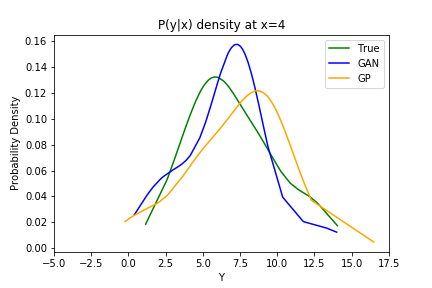}
            \caption{}
            \label{fig:c}
    \end{subfigure}

\end{minipage}
  \begin{minipage}[b]{0.35\textwidth}
\centering
  \begin{tabular}{|c|c|c|}
 \hline
{Method} & {NLPD} & {MAE} \\ \hline
 \hline
 GP (RBF) & 2.539 &  2.461\\\hline
 GP (Quad) & 2.514 & 2.428\\\hline
 DNN & 2.570 & 2.551\\\hline
 XGBoost & 2.947 & 3.262\\\hline
 MDN & 2.529 & 3.176\\\hline
 CGAN & 2.595 &  2.489\\\hline
 \end{tabular}
    \label{table:d1}
    \captionof{table}{NLPD and MAE values on $\texttt{linear}$}
\end{minipage}
\end{minipage}
\caption{\texttt{linear} dataset. $y$ samples generated by: (a) True model, (b) GP (RBF) predictions, and (c) CGAN predictions for various $x$ test values. Subplots (d) and (e) show the three probability densities $p(y|x)$ at $x=0$ and $x=4$.}
\label{fig:d1}
    \end{figure*}

\begin{figure*}[t!]
    \centering
    \begin{subfigure}[b]{0.3\textwidth}
           \centering
           \includegraphics[width=\textwidth]{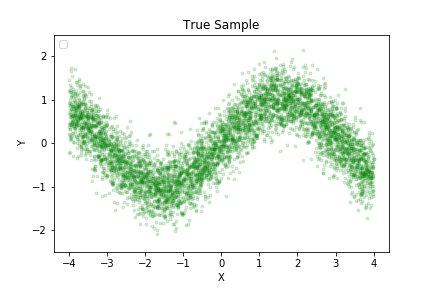}
            \caption{}
            \label{fig:a}
    \end{subfigure}
    \begin{subfigure}[b]{0.3\textwidth}
            \centering
            \includegraphics[width=\textwidth]{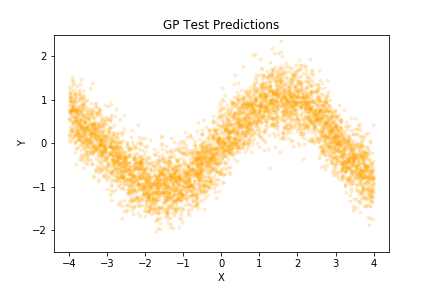}
            \caption{}
            \label{fig:b}
    \end{subfigure}
\begin{subfigure}[b]{0.3\textwidth}
            \centering
            \includegraphics[width=\textwidth]{images/sc1true_more_sample_ypred.png}
            \caption{}
            \label{fig:c}
    \end{subfigure}
\newline\vspace{-1em}
    
      \begin{minipage}{\textwidth}
 \begin{minipage}[b]{0.6\textwidth}
    \begin{subfigure}[b]{0.48\textwidth}
            \centering
            \includegraphics[width=\textwidth]{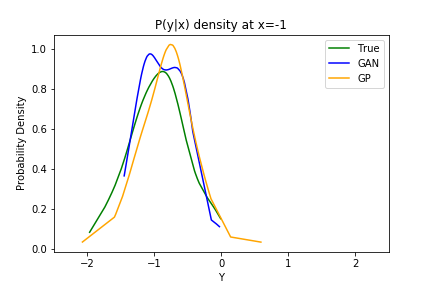}
            \caption{}
            \label{fig:b}
    \end{subfigure}
\begin{subfigure}[b]{0.48\textwidth}
            \centering
            \includegraphics[width=\textwidth]{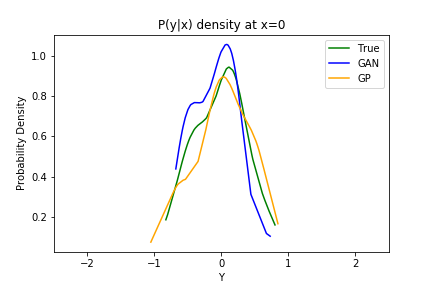}
            \caption{}
            \label{fig:c}
    \end{subfigure}

\end{minipage}
  \begin{minipage}[b]{0.4\textwidth}
\centering
 \begin{tabular}{|c|c|c|}
 \hline
{Method} & {NLPD} & {MAE} \\ \hline
 \hline
 GP (RBF) & 0.553 &  0.339\\\hline
 GP (Quad) & 0.553 & 0.339\\\hline
 DNN & 0.625 & 0.364\\\hline
 XGBoost & 1.141 & 0.459\\\hline
  MDN & 0.541 & 0.467\\\hline
 CGAN & 0.588 &  0.343\\\hline
 \end{tabular}
   \label{table:d3}
    \captionof{table}{NLPD and MAE values on $\texttt{sinus}$}
\end{minipage}
\end{minipage}
    \caption{\texttt{sinus} dataset. $y$ samples generated by: (a) True model, (b) GP (RBF) predictions, and (c) CGAN predictions for various $x$ test values. Subplots (d) and (e) show the three probability densities $p(y|x)$ at $x=-1$ and $x=0$.}
    \label{fig:d2}
    \end{figure*}
    
\begin{figure*}[b!]
    \centering
    \begin{subfigure}[b]{0.3\textwidth}
           \centering
           \includegraphics[width=\textwidth]{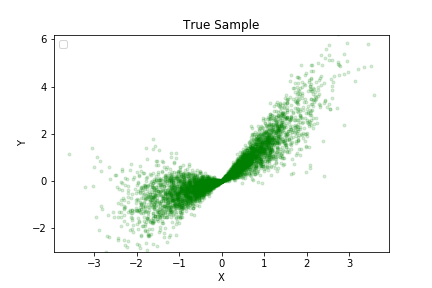}
            \caption{}
            \label{fig:a}
    \end{subfigure}
    \begin{subfigure}[b]{0.3\textwidth}
            \centering
            \includegraphics[width=\textwidth]{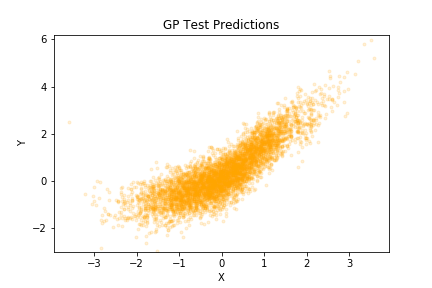}
            \caption{}
            \label{fig:b}
    \end{subfigure}
\begin{subfigure}[b]{0.3\textwidth}
            \centering
            \includegraphics[width=\textwidth]{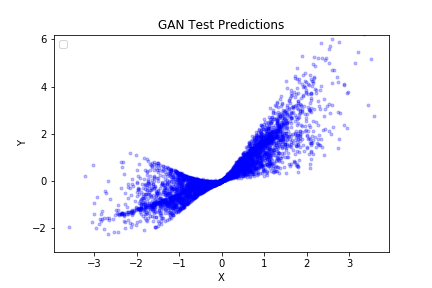}
            \caption{}
            \label{fig:c}
    \end{subfigure}
\newline\vspace{-1em}
    
      \begin{minipage}{\textwidth}
 \begin{minipage}[b]{0.6\textwidth}
    \begin{subfigure}[b]{0.48\textwidth}
            \centering
            \includegraphics[width=\textwidth]{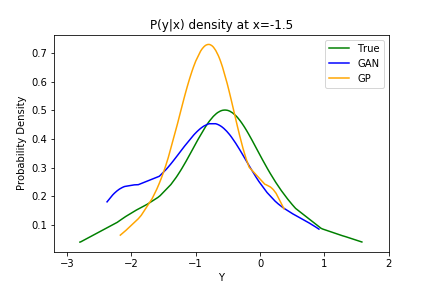}
            \caption{}
            \label{fig:b}
    \end{subfigure}
\begin{subfigure}[b]{0.48\textwidth}
            \centering
            \includegraphics[width=\textwidth]{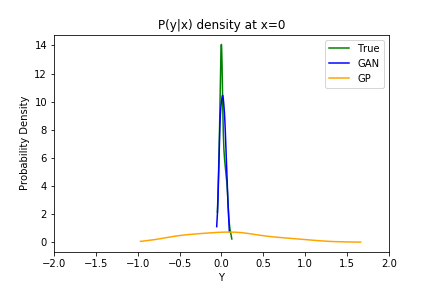}
            \caption{}
            \label{fig:c}
    \end{subfigure}

\end{minipage}
  \begin{minipage}[b]{0.4\textwidth}
\centering
 \begin{tabular}{|c|c|c|}
 \hline
{Method} & {NLPD} & {MAE} \\ \hline
 \hline
 GP (RBF) & 0.732 &  0.333\\\hline
 GP (Quad) & 0.730 & 0.332\\\hline
  DNN & 0.796 & 0.358\\\hline
 XGBoost & 1.512 & 0.449\\\hline
 MDN & \textbf{0.175} & 0.368\\\hline
 CGAN & 0.512 &  0.334\\\hline
 \end{tabular}
   \label{table:d3}
    \captionof{table}{NLPD and MAE values on $\texttt{heteroscedastic}$}
\end{minipage}
\end{minipage}

\caption{\texttt{heteroschedastic} dataset. $y$ samples generated by: (a) True model, (b) GP (RBF) predictions, and (c) CGAN predictions for various $x$ test values. Subplots (d) and (e) show the three probability densities $p(y|x)$ at $x=-1.5$ and $x=0$.}
\label{fig:d3}
    \end{figure*}

\section{Additive noise }
\label{sec:linear}
We first consider the regression scenario with additive noise, $y=f(x)+z$ which is a common assumption made by most regression models. We take $z$ to be Gaussian noise. Commonly used regression models such as GPs, DNNs and XGboost are expected to perform well on it, since they model additive Gaussian noise explicitly. This is a basic test for CGAN, which should also be able to model Gaussian noise easily, though the design happens via implicit modeling.

\paragraph{Gaussian noise dataset ($\texttt{linear}$)}
We first generate a dataset with standard Gaussian noise with $y = x+z$, where $x \sim \mathcal{N}(4,3)$ and $z \sim \mathcal{N}(3,3)$. Figure~\ref{fig:d1} shows the generated samples of the true model, GP and CGAN. Clearly, the sample cloud produced by GP looks more similar to the true samples compared to CGAN. Figure~\ref{fig:d1} also lists NLPD and MAE metric values for all the methods on test data. There is not a significant difference between the regression methods in terms of NLPD and MAE metrics. Since here $f(x) = x$ is quite simple, we next try a complex $f$ to see if CGAN is capable of modeling simple noise with a bit more complex $f$.

\paragraph{Sinusoidal dataset ($\texttt{sinus}$)}
We use Gaussian noise but use $y = sin(x) + z$ to generate a more complex function of $x$, where $z \sim \mathcal{N}(0,1), x \sim \mathcal{U}[-4,4]$ and $\mathcal{U}$ is the uniform distribution. This is again a simple problem for non-linear regression models like GP. Figure~\ref{fig:d2} shows samples generated by these methods and the true function. Here too, GP clearly produces more realistic samples than CGAN. The figure also gives the NLPD and MAE metric values on the test data, with CGAN's performance being a tad lower. This case also shows that CGAN is capable of modeling complex functional relationships between inputs and outputs with a simple linear additive noise form. Even with additive noise form, CGAN has an advantage over GP when the noise is unimodal but exhibits asymmetric tails, \eg exponential noise (refer to the supplementary material for an example). The natural next step is to investigate with non-additive complex noise forms, ones which are generally encountered in real world.

\section{Heteroscedastic noise}
\label{sec:hetero}
The previous datasets use an additive noise form that is independent of $x$. However, real world phenomena exhibit more complex noise. We generate a dataset with heteroschedastic noise ($\texttt{heteroscedastic}$), \ie noise that is dependent on $x$. We use the following generation process: $y = x + h(x, z)$ where $h(x, z) = (0.001 + 0.5 |x|) \times z$ and $z \sim \mathcal{N}(1,1)$. The interesting region of this dataset is around $x = 0$; in this region noise is small compared to other regions of input $x$. Figure~\ref{fig:d3} shows the samples generated by these methods and the true sample distribution. CGAN generates more realistic samples compared to GP that fails to capture the heteroschedastic noise structure, so do other baselines. CGAN also has much better NLPD values as shown in Table 3. While using heteroschedastic GPs~\cite{williams1996gaussian} would also make GPs do better on this dataset, the important point to note here is that, though the CGAN architecture used is similar to the experiments of Section~\ref{sec:linear}, CGAN has easily learnt to model a very different type of noise. This demonstrates the capability of CGAN in modeling complex noise forms. However, MDN performs the best here, as expected since it models noise as a mixture of Gaussians. 

    \begin{figure*}[b!]
    \centering
    \begin{subfigure}[b]{0.3\textwidth}
           \centering
           \includegraphics[width=\textwidth]{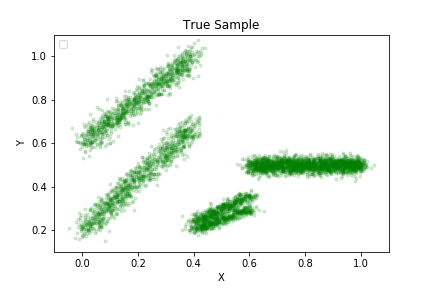}
            \caption{}
            \label{fig:a}
    \end{subfigure}
    \begin{subfigure}[b]{0.3\textwidth}
            \centering
            \includegraphics[width=\textwidth]{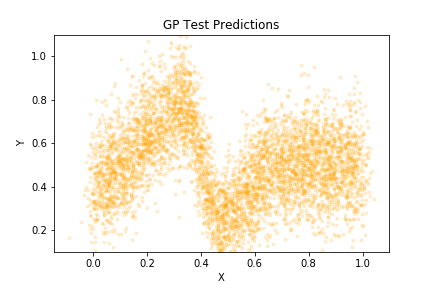}
            \caption{}
            \label{fig:b}
    \end{subfigure}
\begin{subfigure}[b]{0.3\textwidth}
            \centering
            \includegraphics[width=\textwidth]{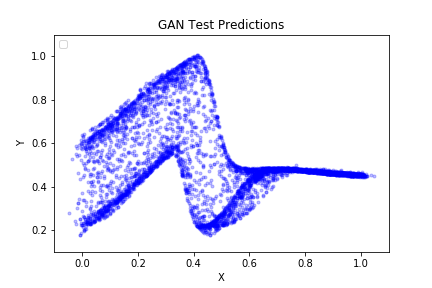}
            \caption{}
            \label{fig:c}
    \end{subfigure}
\newline\vspace{-1em}
    
      \begin{minipage}{\textwidth}
 \begin{minipage}[b]{0.6\textwidth}
    \begin{subfigure}[b]{0.48\textwidth}
            \centering
            \includegraphics[width=\textwidth]{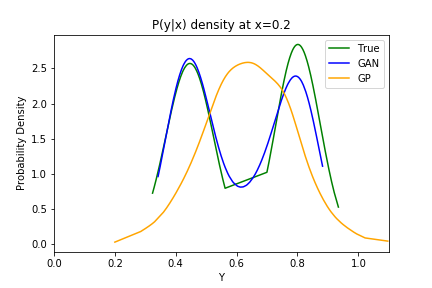}
            \caption{}
            \label{fig:b}
    \end{subfigure}
\begin{subfigure}[b]{0.48\textwidth}
            \centering
            \includegraphics[width=\textwidth]{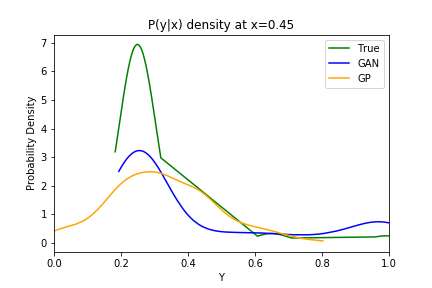}
            \caption{}
            \label{fig:c}
    \end{subfigure}

\end{minipage}
  \begin{minipage}[b]{0.4\textwidth}
\centering
 \begin{tabular}{|c|c|c|}
 \hline
{Method} & {NLPD} & {MAE} \\ \hline
 \hline
 GP (RBF) & -0.600 &  0.095\\\hline
 GP (Quad) & -0.609 & 0.093\\\hline
 DNN & -0.489 & 0.107\\\hline
 XGBoost & -0.164 & 0.106\\\hline
 MDN & \textbf{-1.268} & 0.108\\\hline
 CGAN & \textbf{-1.217} &  0.094\\\hline

 \end{tabular}
   \label{table:d4}
    \captionof{table}{NLPD and MAE values on $\texttt{multi-modal}$}
\end{minipage}
\end{minipage}

\caption{\texttt{multi-modal} dataset. $y$ samples generated by: (a) True model, (b) GP (RBF) predictions, and (c) CGAN predictions for various $x$ test values. Subplots (d) and (e) show the three probability densities $p(y|x)$ at $x=0.2$ and $x=0.45$.}
\label{fig:d4}
    \end{figure*}

\section{Beyond regression: multi-modal posteriors}
\label{sec:multi-response}
Next, we take a case in which the interaction between noise and $x$ is more complex. We construct a simple dataset with a multi-modal $p(y|x)$ distribution ($\texttt{multi-modal}$) which also changes with $x$, to examine whether CGAN can capture such complex distributions. Clearly, the baseline methods are disadvantaged on this task. Such distributions can occur in real world phenomena such as certain dynamical systems which switch between multiple states~\cite{einbeck2006modelling}, depending on latent factors like temperature; this can create scenarios where the same input $x$ can be mapped to two values of $y$. Few models exist in the literature for multi-modal regression~\cite{einbeck2006modelling}. 

We use the following procedure to generate a multi-modal data where $y$ is: $1.2 x + 0.03 z$ or $x + 0.6 + 0.03 z$ when $0.4 < x$; $0.5 x + 0.01 z$ or $0.6 x + 0.01 z$ when $0.4 \leq x < 0.6$ and; $0.5 + 0.02 z$ when $0.6\leq x$, with $z \sim \mathcal{N}(0,1)$ and $x \sim \mathcal{U}[0,1]$.

Figure~\ref{fig:d4} shows the generated test samples and samples predicted by GP and CGAN. CGAN clearly displays a remarkable ability to model the multi-modal nature of the underlying data, while GP expectedly fails to do so since GP can only generate a (unimodal) Gaussian distribution for a given $x$. This also results in poor performance of GP in $x\geq 0.6$ region since it struggles to match the noise across different regions. Few models exist in the literature for multi-modal regression and it is encouraging to see that CGAN can model it implicitly. This demonstrates the versatility of CGAN to capture complex conditional distributions, while other regression methods require special modifications for the same. {\em The above is a demonstration of a powerful feature associated with using CGANs for regression, that the other additive noise models focusing on fitting a central statistic fail at.} However, the MDNs that explicitly model the noise with a mixture of Gaussian likelihood at each $x$, are comparable to CGAN. This is quite important for domains like weather or physical processes where modeling all the possible scenarios is more important than modeling the mean~\cite{hegerl2006climate,kneib2013beyond}. A recent episode with hurricane Dorian's changing path is a demonstration of such phenomenon where modeling possible paths are more important than mean fitting. 

It is useful to note that the ELU activation function smoothens the overall shape of the generated $y$ distribution. Applying sparser activations such as ReLU can help reduce the smoothness (see appendix).  

\begin{figure*}[t]
\centering
    \begin{subfigure}[b]{0.32\textwidth}
           \centering
           \includegraphics[width=\textwidth]{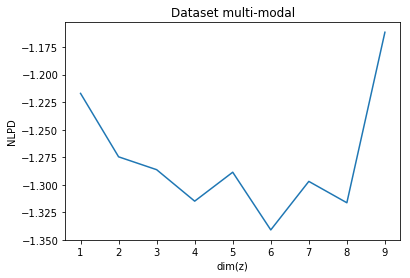}
            \caption{}
            \label{fig:a}
    \end{subfigure}
    \begin{subfigure}[b]{0.295\textwidth}
           \centering
           \includegraphics[width=\textwidth]{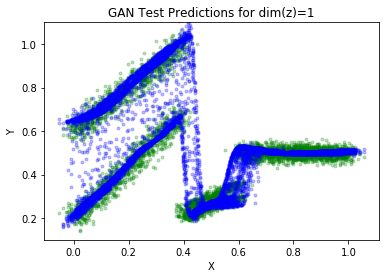}
            \caption{}
            \label{fig:a}
    \end{subfigure}
    \begin{subfigure}[b]{0.295\textwidth}
           \centering
           \includegraphics[width=\textwidth]{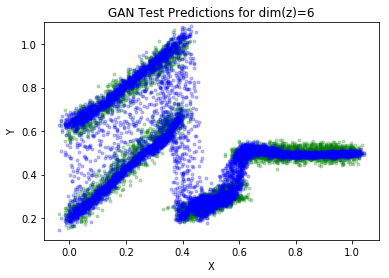}
            \caption{}
            \label{fig:a}
    \end{subfigure}

\caption{Study on input noise dimension $\mathrm{dim}(z)$: (a) shows variation of NLPD with $\mathrm{dim}(z)$ for $\texttt{multi-modal}$; (b) shows the output prediction of CGAN (blue) for $\mathrm{dim}(z)=1$ while (c) shows the output predictions of CGAN (blue) for the best dimension value found in $[1, 10]$, i.e., $\mathrm{dim}(z)=6$ overlayed on true distribution (green). 2 samples were used for each x.}
\label{fig:ablation}
\vspace{-0.5em}
\end{figure*}

\paragraph{Increasing the dimensionality of noise}
To study the effect of dimensionality of noise on CGAN performance, we vary $k=\mathrm{dim}(z)$ over $\{1,2,3, 4,5,6,7,8,9\}$. Figure~\ref{fig:ablation} (a) shows NLPD values as we increase the dimensionality of noise. We get the lowest value of NLPD $=-1.34$ at $k=6$. To point out the goodness of this choice of $\mathrm{dim}(z)$, we show the CGAN prediction samples generated at  $\mathrm{dim}(z)=1$ and $\mathrm{dim}(z)=6$ in Figure \ref{fig:ablation}. There is an improvement in faithfulness to the true distribution density when noise dimensionality is set correctly. 
These results imply that, for problems with complex noise, using a higher dimensional noise can help obtain improved performance. 

   \begin{table*}[b]
  \caption{Comparison of GP - RBF and Quad, DNN, and CGAN on Real World datasets}
\centering
 \scalebox{0.7}{\begin{tabular}{|c||c|c||c|c||c|c||c|c||c|c||c|c||c|c||} 
 \hline
  & \multicolumn{2}{c||}{$\texttt{ailerons}$} & \multicolumn{2}{c||}{$\texttt{CA-housing}$} & \multicolumn{2}{c||}{$\texttt{pumadyn}$} &
  \multicolumn{2}{c||}{$\texttt{bank}$} &
  \multicolumn{2}{c||}{$\texttt{comp-activ}$} &
  \multicolumn{2}{c||}{$\texttt{abalone}$} & \multicolumn{2}{c||}{$\texttt{census-house}$}\\ \hline
 \hline
Method & {NLPD} & {MAE} & {NLPD} & {MAE} & {NLPD} & {MAE}& {NLPD} & {MAE}& {NLPD} & {MAE} & {NLPD} & {MAE} & {NLPD} & {MAE}\\ \hline
 \hline
 GP (RBF) & 1.886 & 1.201 & 0.830 & 0.383 & 4.642 & 19.967 & 1.140 & 0.576 & 2.425 & 2.303 & 2.221 & 1.540 & 0.283 & 0.177 \\
 GP (Quad) & 1.883 & 1.210 & 0.831 & 0.377 &  4.643 & 19.970 & 1.017 & 0.508 & 2.392 & 2.048 & 2.143 & 1.526 & 0.273 & 0.172 \\
 DNN & 2.618 & 1.331 & 1.376 & 0.369 & 4.627 & 7.462 & 1.718 & 0.539 & 4.110 & 2.622 & 2.432 & 1.450  & 0.637 & 0.175\\
 XGboost & 1.945 & 1.214 & 0.665 & 0.306 & 3.516  & 6.499  & 1.072 & 0.508 & 2.259 & 1.643 & 2.269 & 1.666 & 0.287 & 0.167\\
 \hline
 MDN & 1.923 & 1.687 & 0.595  & 0.557  & 4.518 & 27.978 & 1.055 & 0.719  & 2.563 & 4.341 & 1.961 & 2.116 & -0.637 & 0.185\\ 
 \hline
 CGAN & 2.031 & 1.339 & 0.681 & 0.364 & 4.754 & 22.341 & 1.236 & 0.534 & 3.025 &  3.749 & 2.088 & 1.528 & 0.255 & 0.196\\ 
 \hline
 \end{tabular}
 }
   \label{table:rw}
   \vspace{-1em}
 \end{table*}

\section{Real world Datasets}
\label{sec:real-world}
We run experiments on seven real world datasets: ailerons, CA-housing, pumadyn, bank, comp-activ, abalone, and census-house.



Table~\ref{table:rw} depicts the results from these datasets. Overall, XGboost gives the best performance over the MAE metric, while MDN produces the best NLPD values across four datasets (and is close to best on other three). However, there is no method that is a clear winner on all the datasets, though XGBoost clearly is among top-2 across all datasets. MDN prduces inferior MAE metric since due to Gaussian sampling, it has a tendency to produce some absurd outliers that give it higher MAE scores. While \textit{CGAN is competitive with the state-of-the-art methods}, it is not as good as XGBoost or MDN for predictions or modeling noise, respectively. MDN's ability to model noise was also corroborated with synthetic datasets.

\section{Effect of training size}
\begin{figure}[t]
\centering
    \begin{subfigure}[b]{0.23\textwidth}
           \centering
           \includegraphics[width=\textwidth]{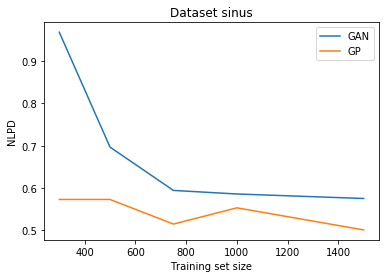}
            \caption{}
            \label{fig:a}
    \end{subfigure}
    \begin{subfigure}[b]{0.23\textwidth}
           \centering
           \includegraphics[width=\textwidth]{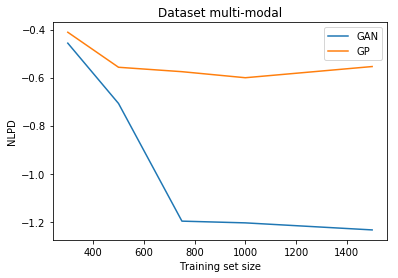}
            \caption{}
            \label{fig:b}
    \end{subfigure}
\caption{The effect of the size of training data for a) $\texttt{sinus}$ b) $\texttt{multi-modal}$. 
}
\label{fig:sc6}
\end{figure}

GP, like other Bayesian methods, is known to perform well with small datasets. This is because these methods average the predictions using the posterior in the weight space. CGAN does not enjoy this property since it uses only a single set of weights. Given this, it is worthwhile to investigate how CGAN behaves compared to GP when the training sample size becomes small. We report the results in Figure \ref{fig:sc6} on \texttt{sinus} and \texttt{multi-modal} as the training size $N$ decreases. Here we take $N$ to be in $\{300, 500, 750, 1000, 1500\}$. We notice that the performance of GP is stable over the range of training set sizes. CGAN suffers from a reduced training set size. Interestingly, on the \texttt{multi-modal} dataset we find that below 300 instances, GP is performing as well as CGAN. This is because, with very small amount of data, CGAN is unable to mdel the multi-modal noise well. 

	\section{Conclusion}
	\label{sec:conc}

The main points of the paper are: (a) the interpretation of CGAN in the context of regression as a great way to model general forms of noise, e.g., $p(y|x)$ is multi-modal in $y$; and, (b) the pointing out that even a straight-forward implementation CGAN can be competitive with a state-of-the-art methods.
There are additional advantages of CGAN: it models noise more naturally; and, it can handle different types of regression models in one formulation and implementation. Also, compared to CGAN for large structured domains such as images, it is much simpler to set up and design because: (a) it does not need any expert level setting up of the generating function, $f$; (b) the noise dimension, $k$ can be chosen to be equal or higher than $m$, the dimension of the output, $y$, and so, "zero" gradient issues are easily avoided; and, (c) since regression problems have either one or at most a few modes in $y$ for a given $x$, "missed modes" or "mode collapse" issues are not worrisome. Therefore, CGAN is worthy of further exploration for solving regression problems. 

However, as we showed across real and synthetic world experiments, CGANs have a long way to go, with methods like MDN and XGBoost having a superior ability to model the noise as well as do predictions. \emph{Another point to note is that while CGANs required huge amounts of hyper-parameter training owing to adversary's complexity, MDN and XGBoost were relatively straightforward to tune - requiring only a fraction of effort.} MDN can model complex noise forms just like CGAN as shown in multi-modal example, with an explicit likelihood training unlike implicit form in CGAN.

 In this paper we only experimented with datasets with one dimensional output, $m=\mathrm{dim}(y)=1$. It would be useful to evaluate CGAN on datasets with multiple, nonlinearly correlated outputs. Problems in joint, multi-time-series prediction, for example those arising in applications such as financial and weather forecasting that involve many unknown variables, can gain a lot from the application of CGAN. 

There are also many generic directions for future work. In this paper, we employed the standard CGAN discriminator that corresponds to minimizing the Jenson-Shannon divergence. Since maximum likelihood corresponds to minimizing KL divergence, f-GAN ideas in~\cite{nowozin2016f} can be used to change the discriminator suitably, and this could result in better NLPD generalization. Design of scalable CGAN implementations suited for regression is another key direction. Designing a Bayesian version of CGAN~\cite{saatci2017bayesian} would give great gains on small training datasets. With these leading to much better performing CGAN implementations, we can return to do a more thorough benchmarking of CGAN against the current best regression solvers.

	
	\bibliographystyle{unsrt}
	\bibliography{nips}
	\newpage
	\appendix
	\section{Using ReLU activation function for \texttt{multi-modal} dataset}
\begin{figure}[b]
\centering
    \begin{subfigure}[b]{0.3\textwidth}
           \centering
           \includegraphics[width=\textwidth]{images/sc3true_more_sample_ypred.png}
            \caption{}
            \label{fig:a}
    \end{subfigure}
    \begin{subfigure}[b]{0.3\textwidth}
            \centering
            \includegraphics[width=\textwidth]{images/sc3gp_more_sample_ypred.png}
            \caption{}
            \label{fig:b}
    \end{subfigure}
\begin{subfigure}[b]{0.3\textwidth}
            \centering
            \includegraphics[width=\textwidth]{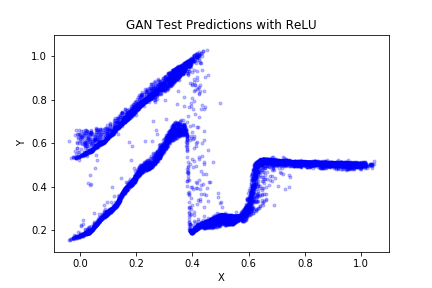}
            \caption{}
            \label{fig:c}
    \end{subfigure}
\newline\vspace{-1em}
    
      \begin{minipage}{\textwidth}
 \begin{minipage}[b]{0.6\textwidth}
    \begin{subfigure}[b]{0.48\textwidth}
            \centering
            \includegraphics[width=\textwidth]{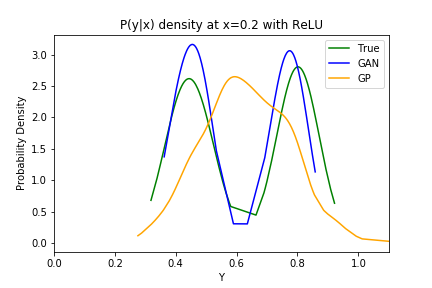}
            \caption{}
            \label{fig:b}
    \end{subfigure}
\begin{subfigure}[b]{0.48\textwidth}
            \centering
            \includegraphics[width=\textwidth]{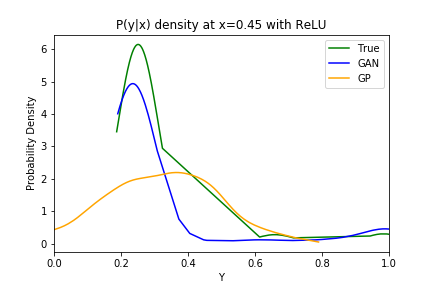}
            \caption{}
            \label{fig:c}
    \end{subfigure}
\end{minipage}
  \begin{minipage}[b]{0.4\textwidth}
\centering
 \begin{tabular}{|c|c|c|}
 \hline
{Method} & {NLPD} & {MAE} \\ \hline
 \hline
 GP (RBF) & -0.600 &  0.095\\\hline
 GP (Quad) & -0.609 & 0.093\\\hline
 DNN & -0.489 & 0.107\\\hline
 XGBoost & -0.164 & 0.106\\\hline
 CGAN (ELU) & \textbf{-1.217} & 0.094\\\hline
 CGAN (ReLU) & -0.903 & 0.105\\\hline
 \end{tabular}
    \captionof{table}{NLPD and MAE values}
\label{table:d4relu}
\end{minipage}
\end{minipage}

\caption{\texttt{multi-modal} dataset using ReLU activation for CGAN. $y$ samples generated by: (a) True model, (b) GP predictions, and (c) CGAN with ReLU predictions for various $x$ test values. Subplots (d) and (e) show the three probability densities $p(y|x)$ at $x=0.2$ and $x=0.45$.}
\label{fig:d4relu}
\end{figure}

In Section 5 in the primary manuscript, we described some experiments on  \texttt{multi-modal} dataset using ELU as activation function. We observe that ELU produces a smoothing over distribution of $y$ when there are discontinuities in $y$ over $x$. See the Figure~\ref{fig:d4relu} (a) below for the discontinuity at $x=0.4$ and $x=0.6$. We perform an experiment to replace ELU with ReLU activation function. Figure~\ref{fig:d4relu} (c) shows the generated samples by this ReLU network. From visual inspection we can see that ReLUs are able to account for these discontinuities better compared to samples obtained via ELU; see also, Figure 4 in the main paper. However, ReLU struggles with forming the noise levels and shapes on the dominant clusters compared to ELU, hence resulting in a lower value of NLPD. In practice we see that ReLU exhibits higher variance on NLPD values compared to ELU across multiple runs since on some runs it struggles to forms the correct noise patterns.

\section{Additive exponential noise (\texttt{exp})}

\begin{figure}
\centering
    \begin{subfigure}[b]{0.3\textwidth}
           \centering
           \includegraphics[width=\textwidth]{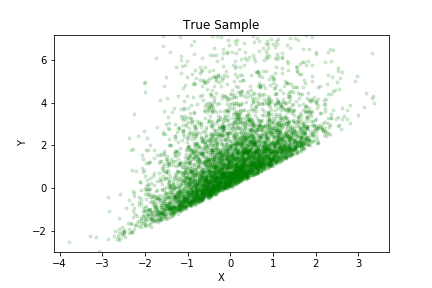}
            \caption{}
            \label{fig:a}
    \end{subfigure}
    \begin{subfigure}[b]{0.3\textwidth}
            \centering
            \includegraphics[width=\textwidth]{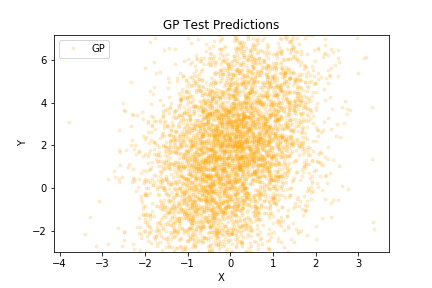}
            \caption{}
            \label{fig:b}
    \end{subfigure}
\begin{subfigure}[b]{0.3\textwidth}
            \centering
            \includegraphics[width=\textwidth]{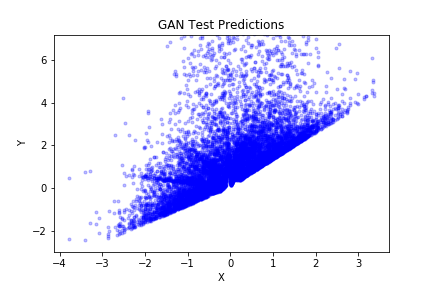}
            \caption{}
            \label{fig:c}
    \end{subfigure}
\newline\vspace{-1em}
    
      \begin{minipage}{\textwidth}
 \begin{minipage}[b]{0.6\textwidth}
    \begin{subfigure}[b]{0.48\textwidth}
            \centering
            \includegraphics[width=\textwidth]{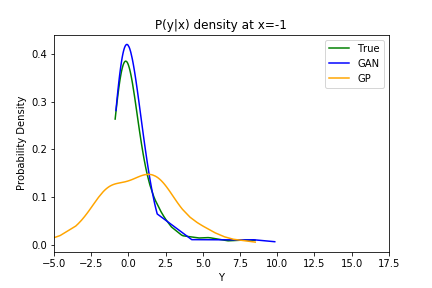}
            \caption{}
            \label{fig:b}
    \end{subfigure}
\begin{subfigure}[b]{0.48\textwidth}
            \centering
            \includegraphics[width=\textwidth]{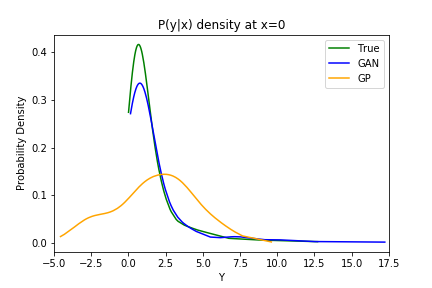}
            \caption{}
            \label{fig:c}
    \end{subfigure}
\end{minipage}
  \begin{minipage}[b]{0.4\textwidth}
\centering
 \begin{tabular}{|c|c|c|}
 \hline
{Method} & {NLPD} & {MAE} \\ \hline
 \hline
 GP (RBF) & 2.131 &  1.334\\\hline
 GP (Quad) & 2.131 & 1.334\\\hline
 DNN & 2.112 &  1.369\\\hline
 XGBoost & 2.585 & 1.829\\\hline
 CGAN & \textbf{2.031} & \textbf{1.205}\\\hline
 \end{tabular}
    \captionof{table}{Comparison of GP and CGAN on $\texttt{exp}$}
\label{table:d4relu}
\end{minipage}
\end{minipage}

\caption{\texttt{exp} dataset. $y$ samples generated by: (a) True model, (b) GP predictions, and (c) CGAN predictions for various $x$ test values. Subplots (d) and (e) show the three probability densities $p(y|x)$ at $x=-1$ and $x=0$.}
\label{fig:dexp}
\end{figure}

As mentioned at the end of Section 4 in the main manuscript, we experiment with CGAN when the noise is unimodal but with asymmetric tails. We use the following generation process for this case: $y = x + \exp(z)$ where $x, z \sim \mathcal{N}(0,1)$. From Figure \ref{fig:dexp} we see that CGAN captures asymmetry while GP assumes normal noise (symmetric); this results in better NLPD values for CGAN. 

\end{document}